\documentclass[10pt,twocolumn,letterpaper]{article}

\usepackage{cvpr}
\usepackage{times}
\usepackage{epsfig}
\usepackage{graphicx}
\usepackage{amsmath}
\usepackage{amssymb}
\usepackage{mathtools}

\usepackage{color}
\usepackage{subfig}
\usepackage{mathabx}
\usepackage{psfrag}
\usepackage{bm}
\usepackage{enumitem}
\usepackage{xcolor}
\usepackage[breaklinks=true,bookmarks=false]{hyperref}
\usepackage{anyfontsize} % to avoid font sizes warnings
\usepackage{soul}

\makeatletter
\renewcommand{\maketag@@@}[1]{\hbox{\m@th\normalsize\normalfont#1}}%
\makeatother

\graphicspath{
  {figs/}
}

\cvprfinalcopy %  Uncomment this line for the final submission

\def\cvprPaperID{3508} %  Enter the CVPR Paper ID here

\ifcvprfinal\fi
\begin{document}

\title{Analytical Modeling of Vanishing Points and Curves in Catadioptric Cameras}

\author{Pedro Miraldo\\
{Instituto Superior T\'{e}cnico, Lisboa}\\
{\tt\small pedro.miraldo@tecnico.ulisboa.pt}
\and
Francisco Eiras\\
{University of Oxford}\\
{\tt\small francisco.eiras@cs.ox.ac.uk}
\and
Srikumar Ramalingam\\
{University of Utah}\\
{\tt\small srikumar@cs.utah.edu}
}

\maketitle
\thispagestyle{empty}

\begin{abstract} 
  Vanishing points and vanishing lines are classical geometrical concepts in perspective cameras that have a lineage dating back to 3 centuries. A vanishing point is a point on the image plane where parallel lines in 3D space appear to converge, whereas a vanishing line passes through 2 or more vanishing points. While such concepts are simple and intuitive in perspective cameras, their counterparts in catadioptric cameras (obtained using mirrors and lenses) are more involved. For example, lines in the 3D space map to higher degree curves in catadioptric cameras. The projection of a set of 3D parallel lines converges on a single point in perspective images, whereas they converge to more than one point in catadioptric cameras. To the best of our knowledge, we are not aware of any systematic development of analytical models for vanishing points and vanishing curves in different types of catadioptric cameras. In this paper, we derive parametric equations for vanishing points and vanishing curves using the calibration parameters, mirror shape coefficients, and direction vectors of parallel lines in 3D space. We show compelling experimental results on vanishing point estimation and absolute pose estimation for a wide range of catadioptric cameras in both simulations and real experiments. 
\end{abstract}
\begin{figure}[t]
  \vspace{-0.3cm}
  \centering
  \subfloat[]{\includegraphics[width=0.230\textwidth]{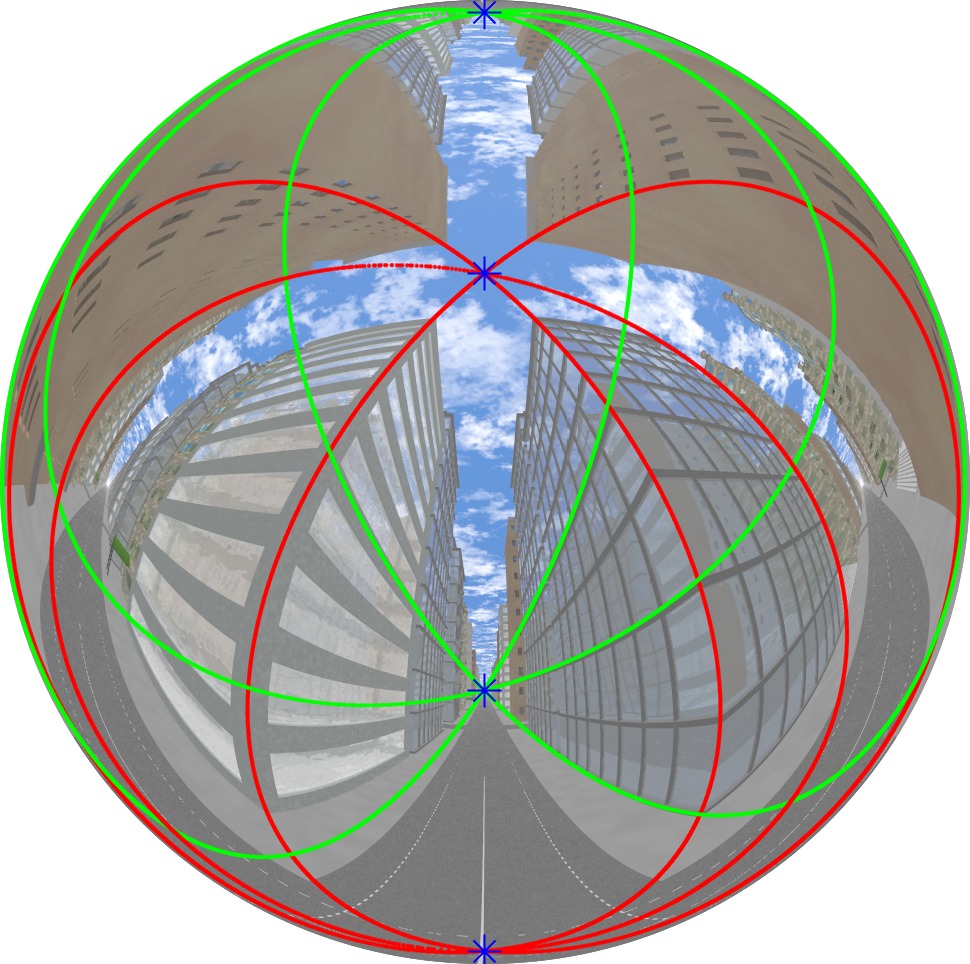}
  \label{fig:example_vanishing_points} }
  \subfloat[]{\includegraphics[width=0.230\textwidth]{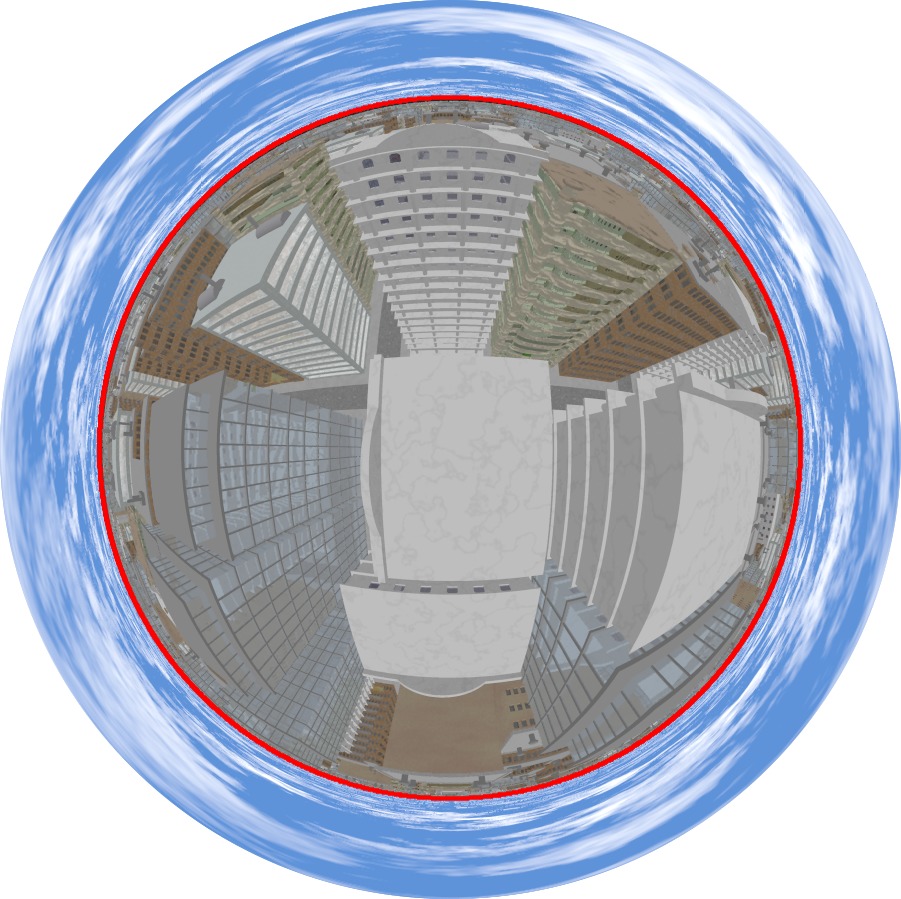}
  \label{fig:example_vanishing_lines} }
  \caption{We illustrate vanishing points and vanishing curves in a catadioptric camera. (a) and (b) show two images from a spherical catadioptric camera. In (a), the red curves denote a set of parallel lines in 3D space, and the two blue points where the red curves converge are the two vanishing points. Similarly we show two other vanishing points associated with a different set of parallel lines that are denoted by green lines. In (b), we show a red vanishing curve passing through two or more vanishing points.
  } 
  \label{fig:example_dir_vanishing_points}
\end{figure}

\section{Introduction}
The idea of vanishing points in catadioptric images is definitely more involved than the classical and well-known image of the point of convergence from parallel railroad tracks. In Fig~\ref{fig:example_dir_vanishing_points}, we show two images of an urban scene captured using a spherical catadioptric camera (a pinhole camera looking at a spherical mirror). We consider two sets of parallel lines and their associated curves in the image space. First, we do not think in terms of straight lines in catadioptric images. Lines in the 3D space are projected to curves in catadioptric images. Furthermore, unlike perspective images, each set of parallel lines converges at more than one vanishing point. For example, in Fig.~\ref{fig:example_dir_vanishing_points}\subref{fig:example_vanishing_points}, the set of parallel lines projects as curves that intersect at two vanishing points. In Fig.~\ref{fig:example_dir_vanishing_points}\subref{fig:example_vanishing_lines}, we show a vanishing curve that passes through more than one vanishing point. In other words, we can think of vanishing curves as some sort of horizon curves. These illustrated vanishing points and vanishing curves can also be expressed in polynomial equations, i.e., expressed 
in parametric equations. The main contribution of this paper is the analytical modeling of vanishing points and vanishing curves for different classes of catadioptric cameras (e.g., spherical, ellipsoidal, or hyperbolic mirrors).

In perspective cameras, vanishing points and vanishing lines are classical concepts that were primarily used by artists for drawings. However, these concepts are not just artistic fascinations, and they have been used by many vision researchers for a wide variety of applications: camera calibration~\cite{caprile90,cipolla99,tardif09,wildenauer12,duan13}, robot control~\cite{schuster93,rives04,ben16}, 3D reconstruction~\cite{parodi96,guillou00,criminisi00,bosse02,ramalingam13} and rotation estimation~\cite{antone00,kosecka02,martins05,denis08,bazin12,camposeco15}.
This paper focuses on catadioptric cameras~\cite{nayar97}, which refers to a camera system where a pinhole camera observes the world from the reflection on a mirror.
While the theory we develop applies to different types of catadioptric cameras, fisheye cameras~\cite{burshardt01} can also be parameterized by catadioptric camera models \cite{ying04,mey07,micusik06}. In particular, our analysis applies to different classes of omnidirectional cameras: central~\cite{baker98,baker99,geyer00,barreto01}, axial, and non-central~ \cite{swaminathan01}. If the cameras satisfy certain conditions and the mirrors belong to specific types (hyperbolic, ellipsoidal, and parabolic), we can achieve a central model. In the case of a spherical mirror, one achieves an axial camera model where all the projection rays pass through a single line in space.

Several methods exist for the extraction of vanishing points without analytical modeling~\cite{tardif09,bazin12,li12,bazin12_2,bazin12_3,xu13,antunes13,antunes17,kluger17}. Vanishing curves are curves in the image that contain a space of solutions for vanishing points, sharing some properties. Similar to vanishing points, these curves have several different applications in computer vision, such as: camera calibration, motion control, geometric reasoning \& image segmentation, and attitude estimation (see for example~\cite{wang90,wang91,okada03,lee09,shabayek12}). In this paper we derive parametric equations for modeling vanishing points and vanishing curves in the case of general catadioptric cameras that can be central, axial, or non-central. To the best of our knowledge, we are not aware of any other work that achieves this. 

When considering the estimation of vanishing points, one needs to take into account the projection and fitting of 3D lines into their images. Several authors studied this problem in central perspective cameras. For central catadioptric cameras, this was studied by \cite{geyer:00,Barreto:01}, in which they parametrize the respective projection curve by a 2\textsuperscript{nd} degree curve. There have been a few results on the projection of lines for non-central catadioptric cameras: 1) in \cite{agrawal10}, Agrawal et~al. defined an analytical solution for the projection curve when considering spherical catadioptric cameras (a 4\textsuperscript{th} degree curve); and 2) in \cite{cameo14}, Cameo et~al. addressed the projection and fitting of lines for cone catadioptric cameras and the camera aligned with the camera's axis of symmetry, in which they also achieved a 4\textsuperscript{th} degree curve. There are alternative solutions to the fitting without implicitly considering the projection of 3D lines, for example: local approximations in small windows by considering the general linear camera approximations \cite{ding09}; or by considering basis functions and look-up tables \cite{yang14}.

While this paper studies the analytical models for vanishing points and vanishing curves, we can think of simpler alternative solutions. For example, the easiest approach would be to approximate the underlying catadioptric model with a central one, correct the distortion, and treat the image as a perspective one, on which one can use classical methods for vanishing point estimation. While this method may work in practice, this approach is not theoretically correct for most catadioptric cameras. A slightly different approach is to reconstruct the 3D lines from 4 or more points on the projected curves~\cite{caglioti05,lanman06,caglioti07,gasparini11,cameo17} in the non-central camera. It is important to note that one can obtain the 3D reconstruction of a line from a single image in a non-central camera. Once we reconstruct the parallel 3D lines in space, we can compute the vanishing point from their intersection. While this method is theoretically elegant, the approach is too brittle in practice, as noticed in \cite{nuno10,agrawal10,agrawal11}.

\subsection{Problem Definition and Contributions} \label{prob:def}
This paper addresses the parameterization of both vanishing points and
vanishing curves, for general (central or non-central) omnidirectional cameras. Our goal is to find parameterization for both, using the following
definitions:
\begin{description}
  \item[Vanishing Points (Sec.~\ref{sec:vanishing_points}):]{
      A common image point that is the intersection of the projection of all
      parallel 3D straight lines (i.e. point at the infinity) onto a camera's
      image, as shown in Fig.~\ref{fig:example_dir_vanishing_points}\subref{fig:example_vanishing_points}. It should, then, only depend on the direction parameters of a 3D
      straight line. 
    }
  \item[Vanishing Curves (Sec.~\ref{sec:curves_infinity}):]{
      The curve that includes vanishing points from straight lines
      perpendicular to some 3D plane in the world, as shown in Fig.~\ref{fig:example_dir_vanishing_points}\subref{fig:example_vanishing_lines}.
    }
\end{description}

We start by considering the projection of points on a
3D straight line, and the case where this point is in the infinity.
With this approach, we are able to extract a polynomial equation that represents
vanishing points as a function of the line's direction. Then, we address the inverse problem, i.e. computing
lines' direction for a given vanishing point.
After defining the vanishing points, we use that parameterization to define the
vanishing curve. 

In addition, to motivate the use of the proposed formulations, we apply the
proposed methods in the relative and the absolute pose problems. The methods
are evaluated and validated using synthetic and real images.

\subsection{Notations} \label{sec:notation}
In our derivation, we will introduce many intermediate polynomial equations. 
Let $\kappa_i^j\left[\, . \, \right]$ denote the $i$\textsuperscript{th} polynomial equation with degree $j$. We will consider a parametric point on the surface of an axially symmetric quadric mirror ($\mathbf{r}\in\mathbb{R}^{3}$), which satisfies the following constraint:
\begin{equation}
  \label{eq:mirror_equation}
  \Omega\left(\mathbf{r}\right) \coloneqq r_1^2 + r_2^2 + Ar_3^2 + Br_3 - C = 0,
\end{equation}
where $A$, $B$, and $C$ are the mirror's parameters. The normal to the mirror at the point $\mathbf{r}$ is given by:
\begin{equation}
  \label{eq:normal_vector_mirror}
  \mathbf{n} = \nabla \Omega(\mathbf{r}) \coloneqq
  \begin{bmatrix}
    r_1 & r_2 & A r_3 + B/2 
  \end{bmatrix}^T.
\end{equation}

To represent 3D lines, by default, we use the notation that represent all
points that belong to that line:
\begin{equation}
  \label{eq:reflectedray}
  \mathbf{p}(\lambda) \coloneqq \lambda \mathbf{s} + \mathbf{q}, \text{  where
  } \lambda \in \mathbb{R}^+.
\end{equation}
Here $\mathbf{q}$ is a 3D point on the line and $\mathbf{s}$ is the direction vector.

\section{Vanishing Points} \label{sec:vanishing_points}
This section addresses the parameterization of vanishing points for general catadioptric camera systems (see Fig.~\ref{fig:example_dir_vanishing_points}\subref{fig:example_vanishing_points}). We observe the projection of a 3D line in space on an image obtained through the reflection on a mirror. Considering a calibrated camera, there is a unique correspondence between a vanishing point on the image and its associated reflection point on the mirror. We will derive a parametric equation that relates reflection points on the mirror, line direction vector, and camera parameters. The parametric equation is beneficial for at least 2 reasons. First, the vanishing points can be estimated from the parametric equation by looking for the projection of points at the infinity on the 3D lines. Additionally, we can also use the parametric equation to extract the direction vector of the parallel lines associated with given vanishing points. 

\subsection{Parameterization of Vanishing Points} \label{sec:computing_vanishing_points}
\begin{figure}[t]
  \subfloat[]{
    \small
    \psfrag{hrid}[c][c]{$\mathbf{r}(\lambda)$}
    \psfrag{n2}[c][c]{$\mathbf{n}(\lambda)$}
    \psfrag{cd}[c][c]{$\mathbf{c}$}
    \psfrag{hpi}[c][c]{$\mathbf{p}(\lambda) = \lambda \mathbf{s} + \mathbf{q} $}
    \psfrag{hdi2}[c][c]{$\mathbf{d}(\lambda)$}
    \includegraphics[width=0.24\textwidth]{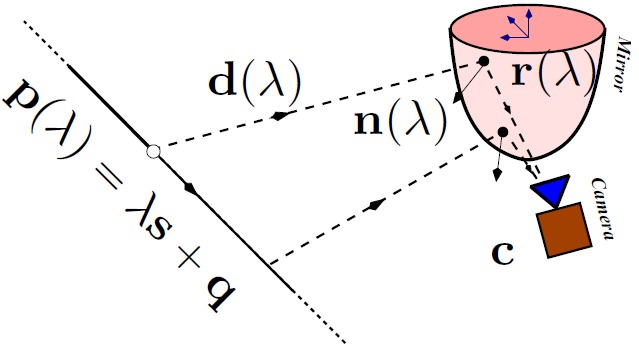}
    \label{fig:problem_def_a}
  }\unskip \vrule 
  \subfloat[]{
    \small
    \psfrag{pi}[c][c]{$\Pi(\lambda)$}
    \psfrag{cd}[c][c]{$\mathbf{c}$}
    \psfrag{hp}[c][c]{$\mathbf{p}(\lambda)$}
    \psfrag{k}[c][c]{$\mathbf{k}(\lambda)$}
    \psfrag{dr}[c][c]{$\mathbf{d}(\lambda)$}
    \psfrag{hdr}[c][c]{$\widetilde{\mathbf{d}}(\lambda)$}
    \includegraphics[width=0.215\textwidth]{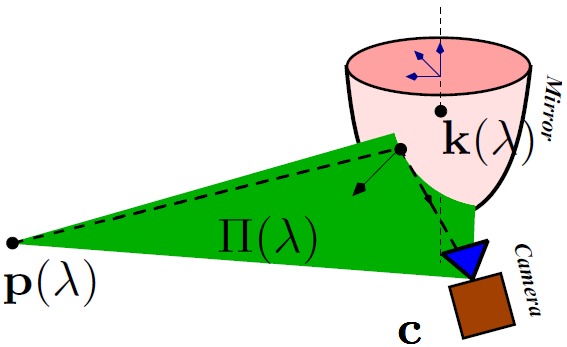}
    \label{fig:problem_def_b}
  }
  \caption{We show the projection of a line in a catadioptric image. The camera does not see the world directly, but observe it through the reflection using the mirror. \protect\subref{fig:problem_def_a} We show the projection of a parameterized point on a line given by $\mathbf{p}(\lambda) \coloneqq \lambda \mathbf{s} + \mathbf{q}$. The point $\mathbf{c}$ and $\mathbf{r}(\lambda)$ are the center of the camera and the reflection point on the mirror, respectively. The normal vector at the reflection point $\mathbf{r}(\lambda)$ is given by $\mathbf{n}(\lambda)$. \protect\subref{fig:problem_def_b} We show the reflection plane that is defined by the camera center ($\mathbf{c}$), reflection point ($\mathbf{r}(\lambda)$) and the parameterized 3D point ($\mathbf{p}(\lambda)$).
  }
  \label{fig:problem_def}
\end{figure}

We briefly provide the general approach and the constraints involved in the parameterization of vanishing points, while the intermediate equations and coefficients are only shown in the supplementary materials. 
Consider Fig~\ref{fig:problem_def}\subref{fig:problem_def_a}, a point on the line $\mathbf{p}(\lambda)$, its reflection point on the mirror $\mathbf{r}(\lambda)$, and the perspective
camera center $\mathbf{c} = [0\ c_2\ c_3 ]$\footnote{Notice that, since we are
 using an axially symmetric quadric mirror, one can rotate the world
coordinate system ensuring that $c_1 = 0$.} define a plane $\Pi(\lambda)$ (denoted as reflection plane), as
shown in Fig.~\ref{fig:problem_def}\subref{fig:problem_def_b}.
Notice that $\mathbf{p}$ and $\mathbf{r}$ depend on the depth $\lambda$ of
the point on the line, while $\mathbf{c}$ is fixed.

Another property of the reflection plane is that it must be parallel to the
normal vector, at the respective reflection point. As a result, any point
$\mathbf{k}(\lambda) = \mathbf{r}(\lambda) + \nu \mathbf{n}(\lambda)$, for any
$\nu$, belongs to the reflection plane. 
Let us consider $\mathbf{r}(\lambda) = \begin{bmatrix} x & y & z
\end{bmatrix}$.  Setting $\nu = -1$, we write:
\begin{equation}
  \mathbf{k}(\lambda) \in \Pi(\lambda) = 
  \begin{bmatrix}
    0 & 0 &  z - A z - B/2
  \end{bmatrix}^T,
\end{equation}
which corresponds to the intersection of the reflection plane $\Pi(\lambda)$ with the $z$--axis of the mirror (mirror's axis of symmetry), as shown in
Fig.~\ref{fig:problem_def}\subref{fig:problem_def_b}. As a result and since $\mathbf{k},\ \mathbf{p}(\lambda), \mathbf{r}(\lambda),\ \mathbf{c} \in \Pi(\lambda)$, one can write:
\begin{equation}
  \text{det}\left(\begin{bmatrix}
      \mathbf{r}(\lambda) & 
      \mathbf{k}(\lambda) & 
      \mathbf{c} & 
      \mathbf{p}(\lambda) \\
      1 & 1 & 1 & 1
  \end{bmatrix}\right) = 0,
\end{equation}
which can be expanded to:
\begin{equation}
  \label{eq:out_plane_def}
  \left( \kappa_1^1\left[z\right] \lambda + \kappa_2^1\left[z\right] \right) x
  + \kappa_3^2\left[y, z\right] \lambda + \kappa_4^2\left[y, z\right] = 0.
\end{equation}
Solving this equation as a function of $x$, we write:
\begin{equation}
  \label{eq:solve_x:mirror_plane}
  x = - \frac{\kappa_3^2\left[y, z\right] \lambda +
  \kappa_4^2\left[y,z\right]}{\kappa_1^1\left[z\right] \lambda +
  \kappa_2^1\left[z\right]}.
\end{equation}

From its definition, vanishing points imply $\lambda = \infty$, in
\eqref{eq:reflectedray}. Then, using \eqref{eq:solve_x:mirror_plane}, we can
derive a constraint:
\begin{equation}
  \label{eq:inf_1}
  x = - \lim_{\lambda \to \infty} \frac{\kappa_3^2\left[y, z\right] \lambda +
  \kappa_4^2\left[y,z\right]}{\kappa_1^1\left[z\right] \lambda +
  \kappa_2^1\left[z\right]}
  = -\frac{\kappa_3^2\left[y, z\right]}{\kappa_1^1\left[z\right]},
\end{equation}
where
\begin{align}
    \kappa_1^1\left[z\right] = & 2s_3c_2 - 2s_2c_3 - Bs_2 + 2s_2(1 - A)z\\
    \kappa_3^2\left[y, z\right] = & s_1(2A - 2)yz + s_1(B + 2c_3)y - 2As_1c_2z \nonumber \\ & -Bs_1c_2.
\end{align}
From the previous equation, one can see that this constraint only depends on
the direction of the line (i.e. parameter $\mathbf{s}$) and the camera
parameters.

Using Snell's law of reflection, we have:
\begin{equation}
  \label{eq:second_constraint}
  \mathbf{d}(\lambda) \times \left( \mathbf{p}(\lambda) - \mathbf{r}(\lambda) \right) = \mathbf{0}.
\end{equation}
We rewrite it as follows:
\begin{equation}
  \label{eq:incident_reflection_ray} 
\mathbf{d}(\lambda) \sim 4 \; \|\mathbf{n}(\lambda)\|\;
\widetilde{\mathbf{d}}(\lambda) - 8 \; \mathbf{n}(\lambda) \langle \widetilde{\mathbf{d}}(\lambda) , \mathbf{n}(\lambda) \rangle,
\end{equation}
where $\widetilde{\mathbf{d}}$ is the reflected ray at the point $\mathbf{r}$.
To conclude, still from Fig.~\ref{fig:problem_def}\subref{fig:problem_def_a},
we define $\widetilde{\mathbf{d}}$ as:
\begin{equation}
  \label{eq:incident_ray}
  \widetilde{\mathbf{d}}(\lambda) \sim \mathbf{r}(\lambda) - \mathbf{c}.
\end{equation}
Replacing \eqref{eq:incident_ray} and \eqref{eq:normal_vector_mirror} in
\eqref{eq:incident_reflection_ray}, and this last one in
\eqref{eq:second_constraint}, we get three equations:
\begin{equation}
  \kappa_5^3[x,y,z,\lambda] = 0,\ 
  \kappa_6^3[x,y,z,\lambda] = 0, \
  \kappa_7^3[y,z,\lambda] = 0 .\label{eq:constraint2_3}
\end{equation}
To ease the calculation and since they are linearly dependent, we choose
$\kappa_7^3[y,z,\lambda]$. This one does not depend on the $x$ variable
and is linear in $\lambda$. Then, one can extract $\lambda$ as: 
\begin{equation}
  \label{eq:solution_lambda}
  \lambda = - \frac{\kappa_8^3\left[y,z\right]}{\kappa_9^3\left[y,z\right]}.
\end{equation}

Again, from its definition, vanishing points imply $\lambda =~\infty$. This
means that $\kappa_9^3[y,z]$ in \eqref{eq:solution_lambda} must be equal to
zero. This equation only depends on the lines direction (i.e. parameters
$\mathbf{s}$). Then, using \eqref{eq:inf_1}, \eqref{eq:solution_lambda}, and
the mirrors equation \eqref{eq:mirror_equation}, one can compute vanishing
points on the mirror for a given direction $\mathbf{s}\in\mathbb{R}^3$, by
solving the system:
\begin{equation}
  \label{eq:vanishing_point}
  \left\{
    \begin{array}{l}
      x \kappa_1^1\left[z\right] = - \kappa_3^2\left[y,
      z\right] \\
      \kappa_9^3\left[y,z\right] = 0 \\  
      x^2 + y^2 + Az^2 + Bz - C = 0,
    \end{array}
  \right.
\end{equation}
where:
\begin{multline}
  \label{eq:kappa_9_new}
  \kappa_{9}^3\left[y, z\right] = a_1y^2 + a_2yz^2+a_3yz + \\
  + a_4y + a_5z^3 + a_6z^2 + a_7z + a_8.
\end{multline}

In the next subsection we present a method to solve this system of
equations.

\subsection{Computing Vanishing Points from a Given Direction}\label{sec:compute_vanishing_points}
To compute the coordinates of a vanishing point on the mirror $\mathbf{r} =
\begin{bmatrix}x & y & z \end{bmatrix}$, for a given direction, in this
subsection we present a method to solve \eqref{eq:vanishing_point}. We start by
replacing $x$ in the first equation, using the mirror's equation, and squaring both sides of the equation:
{\small\begin{equation}
  \label{eq:vanishing_point_1}
  \left\{
    \begin{array}{l}
      \kappa_{10}^4\left[y,z\right] = \kappa_3^2\left[y, z\right]^2 + (y^2 + Az^2 +
      Bz - C)\kappa_1^1\left[z\right]^2= 0,\\
      \kappa_9^3\left[y,z\right] = 0   
    \end{array} 
  \right.
\end{equation}}where:
\begin{multline}
  \label{eq:kappa_17_new}
  \kappa_{10}^4\left[y, z\right] = b_1y^2z^2 + b_2y^2z + b_3y^2 + b_4yz^2 + b_5yz +\\+ b_6y + b_7z^4 + b_8z^3 + b_9z^2 + b_{10}z + b_{11}.
\end{multline}

Now, to solve this problem, since $\kappa_9^3[y,z]$ has degree two in $y$ (see
\eqref{eq:kappa_9_new}), we solve this polynomial equation as a function of
$y$, which gives:
\begin{equation}
  \label{eq:solution_y_vanishing_points} 
  y =  \frac{\kappa_{11}^2[z] \pm \sqrt{\kappa_{12}^4[z]}}{2a_1}, 
\end{equation}
By replacing it in \eqref{eq:kappa_17_new}, we obtain:
\begin{equation}
  \kappa_{13}^2[z]\kappa_{12}^4[z] + \kappa_{14}^4[z] \pm
  \kappa_{15}^2[z]\sqrt{\kappa_{12}^4[z]} = 0.
\end{equation}
After some simplifications, we write a 10\textsuperscript{th} degree polynomial equation as:
{\small\begin{equation}
  (\kappa_{13}^2[z]\kappa_{12}^4[z] + \kappa_{14}^4[z])^2 - 
  \kappa_{15}^2[z]^2\kappa_{12}^4[z] = 0 \Rightarrow
  \kappa_{16}^{10}[z] = 0.
\end{equation}}To compute the coordinates of the reflection of the
vanishing point, one has to compute the solution for $z$ (the real roots
of $\kappa_{16}^{10}[z]$). Then, for each solution of $z$, we get the respective $y$
from \eqref{eq:solution_y_vanishing_points}, and $x$ using the first equation
of \eqref{eq:vanishing_point}.

\begin{table}[t]
  \centering
  \caption{Degrees of the polynomial equation that can be used to compute
  vanishing points, for specific catadioptric camera systems. The * denotes the specific configurations with analytical solutions, which are shown in the supplementary material.}
  \label{tab:poly_degree_compute_points}
  \footnotesize
  \begin{tabular}{|r|c|}
    \hline
    \multicolumn{1}{|r|}{\textbf{Mirror Type}} & \textbf{Degree of $\kappa_{16}[z]$} \\ \hline \hline
    General                              & 10            \\ \hline
    General Axial ($c_2 = 0$)            & 8             \\ \hline
    Spherical Axial ($A = 1$, $B = 0$, $c_2 = 0$)   & 4             \\ \hline
    Ellipsoid Axial ($A = 0$, $C = 0$, $c_2 = 0$)   & 6             \\ \hline
    Conical Axial ($B = 0$, $C = 0$, $c_2 = 0$)     & 4 and $z = 0$ \\ \hline
    Cylindrical Axial ($A = 0$, $B = 0$, $c_2 = 0$) & 4             \\ \hline
    Central with Ellipsoidal Mirror      & 4*             \\ \hline
    Central with Hyperboloidal Mirror    & 4*             \\ \hline
  \end{tabular}
\end{table}

When considering specific types of camera positions (e.g. axial catadioptric
systems) and the use of specific types of mirrors, the polynomial equation
$\kappa_{16}^{10}[z]$ becomes simpler, in terms of its coefficients and its
final polynomial degree. Results for several types of configurations are shown in Tab.~\ref{tab:poly_degree_compute_points}. Furthermore, we observed that the use of unified central model~\cite{geyer00,barreto01} simplified the theory for the central case. The projection of lines using the unified central model \cite{geyer00,barreto01} is given by a 2\textsuperscript{nd} degree polynomial, which makes the relation between 3D lines and vanishing points simpler. For this case, we derive a 4\textsuperscript{th} degree polynomial equation that relates a line's direction with vanishing points, which is shown in the supplementary material.

In the next subsection, we present an easy method to extract directions from
vanishing points.

\subsection{Getting Direction from Vanishing Points}\label{sec:compute_direction_vanishing_points}
Now, let us consider that we have a vanishing point given by $\mathbf{r} = [x,
y, z]$. This subsection addresses the estimation
the direction that yields the respective vanishing point. For
this purpose, we rewrite the system of \eqref{eq:vanishing_point} as a function
of $\mathbf{s}$, which gives:
\begin{equation}
  \label{eq:get_directions}
  \left\{
    \begin{array}{l}
      \kappa_{17}^1\left[s_2,s_3\right] = a_1s_2 + a_2s_3 = 0 \\ 
      \kappa_{18}^2\left[s_1, s_2, s_3 \right] = b_1s_1^2 + b_2s_2^2 +
      b_3s_2s_3 + b_4s_3^2 = 0\\
      \kappa_{19}^2\left[s_1,s_2,s_3\right] = s_1^2 + s_2^2 + s_3^2 - 1 = 0 .
    \end{array} 
  \right.
\end{equation}

To get the direction from a given vanishing points, one needs to solve the system of \eqref{eq:get_directions}. First, we compute $s_3$ from $\kappa_{17}^1[s_2,s_3]$, and replace it in the second and third equations of the system of \eqref{eq:get_directions}, which gives:
\begin{equation}
  \label{eq:get_directions2}
  \left\{ \begin{array}{l} 
      b_1s_1^2 + \left(b_2 -
      b_3\frac{a_1}{a_2} + b_4\frac{a_1^2}{a_2^2}\right)s_2^2 = 0\\
      s_1^2 + \left(1 +
      \frac{a_1^2}{a_2^2}\right)s_2^2 - 1 = 0 .
  \end{array} \right.
\end{equation}
Through elimination, we obtain the solution for $s_2$, as:
\begin{equation}
  \label{eq:solving_s_2}
  s_2^2 = \frac{-b_1}{\left(b_2 - b_3\tfrac{a_1}{a_2} +
  b_4\tfrac{a_1^2}{a_2^2} - b_1 - b_1\tfrac{a_1^2}{a_2^2}\right)}
\end{equation}

To conclude, to obtain the coordinates of the direction from the vanishing
point $\mathbf{r}$, one can compute $s_2$ from \eqref{eq:solving_s_2}, obtain
$s_3 = \tfrac{-a_1 s_2}{a_2}$, and $s_1$ from $
s_1 = \pm \sqrt{1 - s_2^2 - s_3^2}$.

\section{Curves at the Infinity} \label{sec:curves_infinity}
A vanishing curve can be seen as a curve that contains all possible vanishing points associated with a 3D ground plan (a plan of parallel lines). In this section, we will derive an equation to represent this curve in the mirror's surface, as a function of the vector that is normal to a 3D plane.

As presented in Sec.~\ref{prob:def}, vanishing curves are defined by the set of
vanishing points that result from directions perpendicular to some normal
vector $\mathbf{n}$. To start our derivations, we define $\mathbf{s}_1$ and
$\mathbf{s}_2$ as the basis for the space of directions perpendicular to
$\mathbf{n}$ as:
\begin{equation}
  \label{eq:plane_def}
  \mathbf{s} = \alpha \mathbf{s}_1 + (1 - \alpha) \mathbf{s}_2,
\end{equation}
such that $\mathbf{s}^T\mathbf{n} = 0$ for any $\alpha\in\mathbb{R}$. For that
purpose, let us consider three possible solutions for $\mathbf{s}_{1}$ and
$\mathbf{s}_2$, as $\left\{ \mathbf{s}^+, \mathbf{s}^-, \mathbf{s}^* \right\}$:
\begin{align}
  \mathbf{s}^+ = &\ \mathbf{n}\times \begin{bmatrix} 1 & 0 & 0 \end{bmatrix} =
  \begin{bmatrix} 0 & n_3 & -n_2 \end{bmatrix}\\
  \mathbf{s}^- = &\ \mathbf{n}\times \begin{bmatrix} 0 & 1 & 0 \end{bmatrix} =
  \begin{bmatrix} -n_3 & 0 & n_1 \end{bmatrix}\\
  \mathbf{s}^* = &\ \mathbf{n}\times \begin{bmatrix} 0 & 0 & 1 \end{bmatrix} =
  \begin{bmatrix} n_2 & -n_1 & 0 \end{bmatrix}.
\end{align}
Each of the three vectors is either null or orthogonal to $\mathbf{n}$, with at least two nonzero ones. We associate to $\mathbf{s}_1$ and $\mathbf{s}_2$ the two
vectors in the $\{\mathbf{s}^+, \mathbf{s}^-, \mathbf{s}^*\}$ with the highest
norm.

Assuming that both $\mathbf{s}_1$ and $\mathbf{s}_2$ are defined as presented
above, we use the parameterization of the direction \eqref{eq:plane_def} in the
constraint \eqref{eq:inf_1}.
After some simplifications, one can define a 3D surface $\Gamma(\mathbf{r})$ as:
\begin{equation}
  \Gamma(\mathbf{r}) \coloneqq \kappa_{20}^2\left[y, z\right] x + \kappa_{21}^3[y, z] = 0,
\end{equation}
The coefficients for this polynomial are defined in the supplementary material.

To conclude, one can define vanishing curves by the intersection of
$\Gamma(\mathbf{r})$ with the quadric mirror, such that:
\begin{equation}
  \label{eq:solving_curve_infinity}
  \gamma (\mathbf{r}) \coloneqq \left\{ \mathbf{r} = [x,y,z] \in \mathbb{R}^3:
  \Gamma(\mathbf{r}) \wedge \Omega(\mathbf{r}) = 0 \right\}.
\end{equation}
Supplementary material provides a method to solve \eqref{eq:solving_curve_infinity}.

\section{Applications and Experimental Results}
We performed experiments using simulation and real data.
We evaluate the proposed solutions: (1) the computation of vanishing points from a given direction; and (2) to recover the directions from vanishing points (Sec.~\ref{sec:experiments_noise_evaluation_vp}). We present a method to compute the camera's pose using vanishing points (Sec.~\ref{sec:app_camera_pose}) and present some information regarding the projection of 3D lines onto the image. To conclude, we exploit the relative pose problem (Sec.~\ref{sec:relative_pose}).

Real data was used to estimate both vanishing points and vanishing curves. We show some results for the estimation of the absolute camera pose (Sec.~\ref{sec:real_data}), modeling of central and non-central systems (Sec~\ref{sec:central_vs_ncentral}), and for the parameterization of the vanishing curve (Sec.~\ref{sec:exp_vanishing_lines}).

\begin{figure}
  \vspace{-0.3cm}
  \centering
  \subfloat[]{\includegraphics[height=3.55cm]{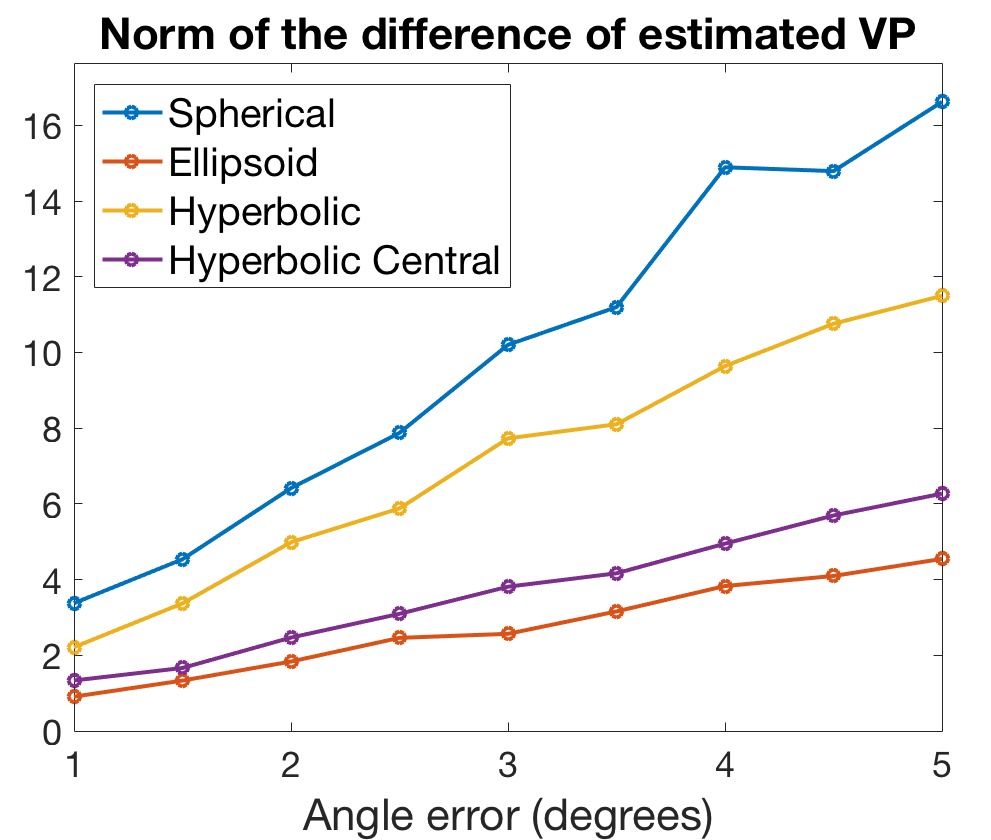}\label{fig:error_vp}}
  \subfloat[]{\includegraphics[height=3.55cm]{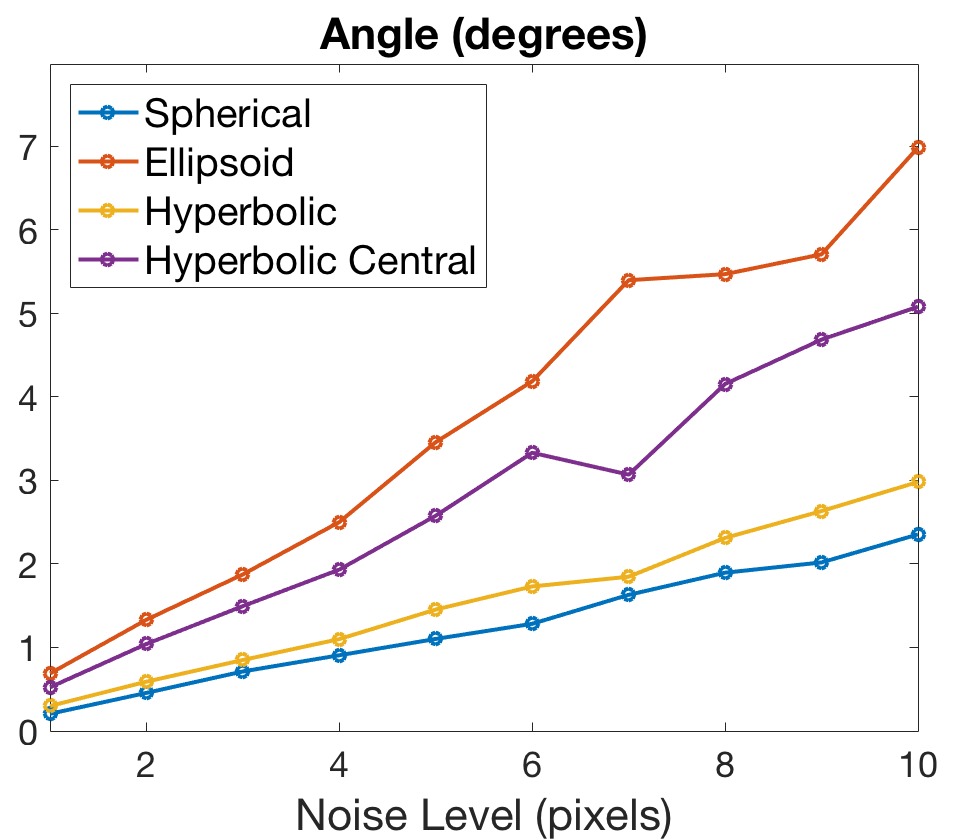}\label{fig:error_dir}}
  \caption{(a) Computation of vanishing points from line directions \protect\subref{fig:error_vp}. (b) Estimation of the direction from vanishing points \protect\subref{fig:error_dir}. We consider four different camera configurations under different noise settings. 
  }
  \label{fig:error}
\end{figure}

\subsection{Evaluation of the Vanishing Points and the Direction Estimation with Noise}\label{sec:experiments_noise_evaluation_vp}

In this subsection we evaluate the estimation of vanishing points in the presence of noise in simulation data. For a given direction $\mathbf{s}_{\text{GT}}$, we compute the ground truth vanishing point $\mathbf{r}_{\text{GT}}$ as described in Sec.~\ref{sec:compute_vanishing_points}. The effect of noise in the data is simulated by changing the direction vector $\mathbf{s}$ by a random angle. For that direction, we compute the vanishing point $\mathbf{r}$, and measure the distance error: $\delta_i =
\|\mathbf{r}_{\text{GT}}-\mathbf{r}_i\|$. This procedure is repeated varying the noise level from $0$ to $5$ degrees, and for each level of
noise we consider $10^2$ randomly generated trials. Results for four different
systems are shown in Fig.~\ref{fig:error}\subref{fig:error_vp}. Although they vary significantly for the considered configurations, the errors are (approximately) linearly with the noise for all of them. The configurations that use ellipsoid and central hyperbolic mirrors show betters results than the spherical and hyperbolic \footnote{Note that these results also depend on the camera position w.r.t. the mirror, which we keep constant for all the experiments with the exception of the hyperbolic central that requires a specific camera position.}.

To evaluate the computation of a direction associated with a given vanishing point, we: 1) obtain the vanishing point for a given direction $\mathbf{s}_{\text{GT}}$,
Sec.~\ref{sec:computing_vanishing_points}; 2) add noise to the coordinates of
the computed vanishing points (a normal distribution with a variable standard
deviation, denoted as {\tt Noise\_Level} and a mean of $0$); 3) compute the
direction $\mathbf{s}$, as shown in
Sec.~\ref{sec:compute_direction_vanishing_points}; and 4) measure the angle
between the estimated direction and the ground truth.
This procedure is repeated $10^2$ times for each level of noise (from $0$ to
$10$ pixels), and the results are shown in Fig.~\ref{fig:error}\subref{fig:error_dir}.
Similar to what happens in the results presented in Fig.~\ref{fig:error}\subref{fig:error_vp}, in this case the errors also vary linearly with the noise for all of the configurations, with configurations with hyperbolic central and ellipsoidal mirrors performing worse than the rest.

\begin{figure}
  \vspace{-0.3cm}
  \centering
  \subfloat[]{\includegraphics[height=3.45cm]{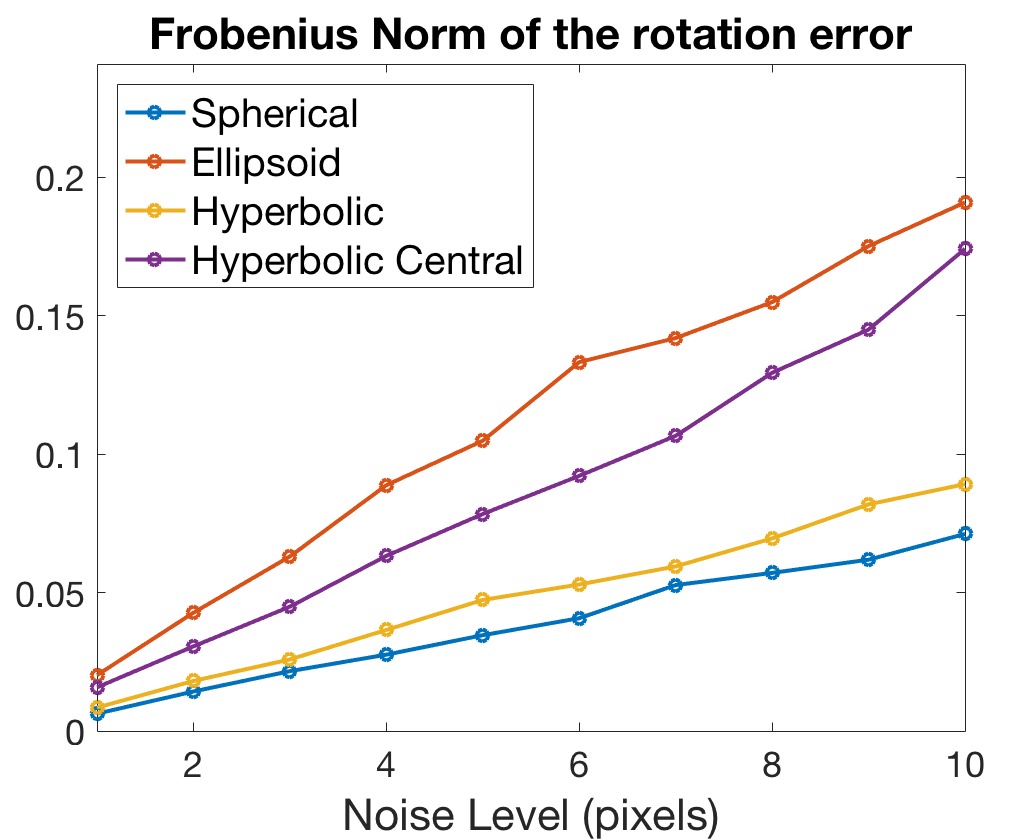}\label{fig:error_rotation}}
  \subfloat[]{\includegraphics[height=3.45cm]{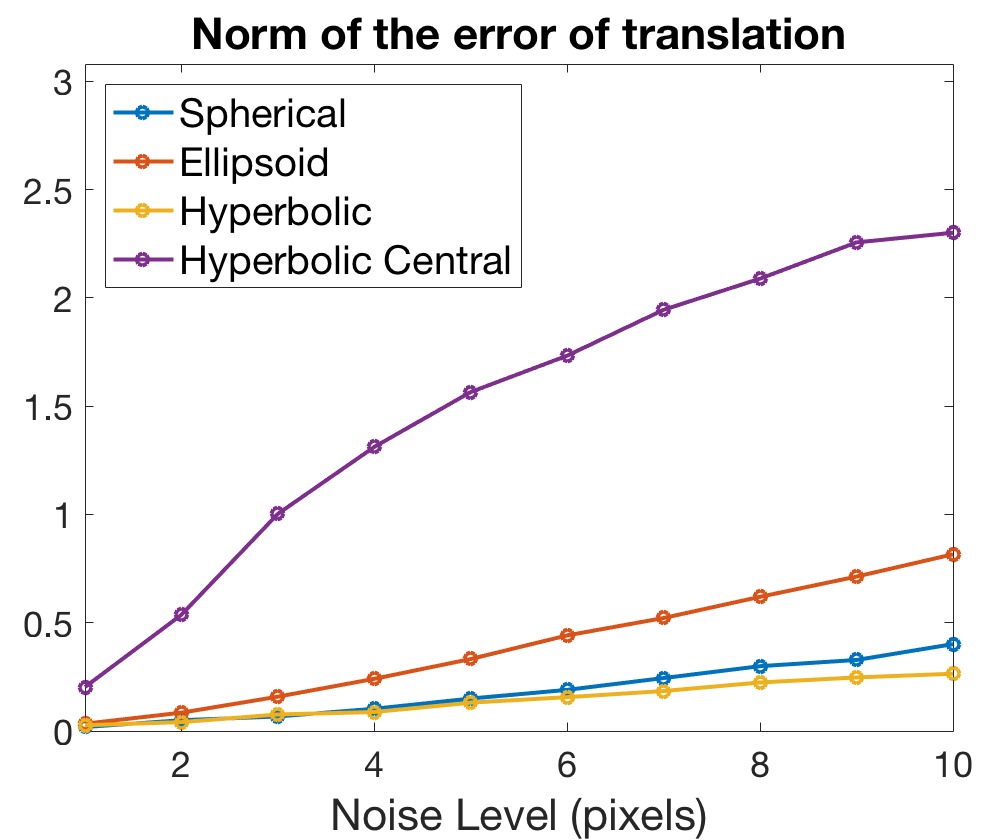}\label{fig:error_translation}}
  \caption{Evaluation of the camera pose using vanishing points, as a function of the noise level in the image. \protect\subref{fig:error_rotation} shows the results for the estimation of the rotation while, \protect\subref{fig:error_translation} presents the results for the translation parameters. For that purpose, we consider four different camera configurations under different noise settings. 
  }
  \label{fig:cam_pose}
\end{figure}

\subsection{Camera Pose}\label{sec:app_camera_pose}
This subsection addresses the estimation of the absolute camera pose, from two or more vanishing points. 

\subsubsection{Estimation of the Rotation Parameters} \label{sec:experiments_estimation_rotation_parameters}

From the results presented above, using $N \geq 2$ vanishing points, we can
recover $N$ different directions $\{\mathbf{s}_{1}, \dots,
\mathbf{s}_{N}\}$. Knowing these directions in the world coordinate system
(let us denote these as $\{\widecheck{\mathbf{s}}_1, \dots,
\widecheck{\mathbf{s}}_N\}$), we use the orthogonal Procrustes' problem
\cite{schonemann66} to compute the rotation $\mathbf{R}\in\mathcal{SO}(3)$ that
aligns camera and world coordinate systems.

To evaluate the estimation of the rotation parameters in an absolute camera
pose application, we consider the following procedure: 1) we selected
three random directions in the world; 2) apply a {\it ground truth}
rotation $\mathbf{R}_{\text{GT}}\in\mathcal{SO}(3)$ to these directions; 3)
compute the respective vanishing points; 4) add noise to the vanishing points
using the variable {\tt Noise\_Level} (similar to what was done in
Sec.~\ref{sec:experiments_noise_evaluation_vp}); 5) compute the 3D direction
from the noisy vanishing points; 6) compute the rotation $\mathbf{R}$
using the Procrustes' problem; and 7) compute the error in the estimation
of the rotation matrix by $\alpha_{\text{Err}} = \| \mathbf{R}_{\text{GT}} - \mathbf{R} 
\|_{\text{Frob}}$. We repeat this procedure $10^2$ for each level of noise for four
different configurations. Results for this evaluation are shown in
Fig.~\ref{fig:cam_pose}\subref{fig:error_rotation}. The errors in the
estimation of the rotation matrix vary linearly with the noise in the image,
and the behaviour in terms of camera configuration is similar to the one in
Fig.~\ref{fig:error}\subref{fig:error_dir} (which makes sense because the rotation
estimation depends on the method used to get directions from vanishing points).

We note that this method can be applied to $N = 2$, by considering $\widecheck{\mathbf{s}}_3 = \widecheck{\mathbf{s}}_1 \times
\widecheck{\mathbf{s}}_2$. An example of the rotation estimation using only two
vanishing points is shown below.

\subsubsection{Estimation of the Translation Parameters}\label{sec:experiments_estimation_translation_parameters}
For the estimation of the pose's translation parameters
$\mathbf{t}\in\mathbb{R}^{3}$, one needs to use more information than vanishing
points. In this subsection, we show a technique that, based on previously
estimated rotation parameters (which was computed from
Sec.~\ref{sec:experiments_estimation_rotation_parameters}) and with knowledge
of the coordinates of (at least) three image points that are images of two 3D
straight lines\footnote{The minimal case corresponds to two image points belonging to a 3D line and one more image point in another 3D line.}, is able to compute the camera's translation parameters.

To simplify the notation, let us now represent the 3D lines using {\it
Pl\"ucker} coordinates~\cite{pottmann01}, by
$\widecheck{\mathbf{l}}_{i}\in\mathbb{R}^{6} = (\widecheck{\mathbf{s}}_{i}, \
\widecheck{\mathbf{m}}_{i})$, for $i=1,\dots, N$ , where
$\widecheck{\mathbf{m}}_{i} = \widecheck{\mathbf{q}}_{i} \times
\widecheck{\mathbf{s}}_{i} $. Now, for a set of pixels that are images of these
3D straight lines, one can map them into 3D straight lines (inverse projection map)
$\mathbf{l}_{i,j}\in\mathbb{R}^{6} = (\mathbf{s}_{i,j}, \ \mathbf{m}_{i,j})$,
where $j = 1, \dots, M_i$ and $M_i$ represents the number of points that are
images of the $i^{\text{th}}$ line. Using the results of
\cite{pless03}, one can write:
\begin{equation}
  \langle [\mathbf{t}]_{\text{x}}\mathbf{R}\widecheck{\mathbf{s}}_i, \mathbf{s}_{i} \rangle +
  \langle \mathbf{R}\widecheck{\mathbf{s}}_i, \mathbf{m}_{i} \rangle +
  \langle \mathbf{R}\widecheck{\mathbf{m}}_i, \mathbf{s}_{i} \rangle = 0.
\end{equation}
Since $\mathbf{R}$ is known, after some calculations, one can rewrite this
constraint as: $\mathbf{a}_{i,j}^T\mathbf{t} = b_{i,j}$\footnote{The elements
of $\mathbf{a}_{i,j}$ and $b_i,$ can be easily obtained after some symbolic
computation.}. Stacking $\mathbf{a}_{i,j}$ and $b_{i,j}$ for a set of
$\{i,j\}$, one can compute the translation parameters as:
\begin{equation}
  \mathbf{A}\mathbf{t} = \mathbf{b} \Rightarrow
  \mathbf{t} = \mathbf{A}^{\dag}\mathbf{b},
\end{equation}
where $\mathbf{A}^{\dag}$ represents the pseudo inverse of $\mathbf{A}$.

To evaluate this approach, we consider the same data-set presented in the
previous subsection and define pixels that are images of two 3D lines.
After obtaining $\mathbf{R}$, we apply the method presented in this subsection. Noise was included in both the vanishing points and the pixels that are images of 3D lines (normal distribution with the standard deviation equals to the {\tt Noise\_Level} variable). Results are shown in Fig.~\ref{fig:cam_pose}\subref{fig:error_translation}\footnote{In these tests, we consider image pixels from two 3D straight lines. If more 3D lines were considered, we would have a more robust estimation.}. From these results, with the exception of the central hyperbolic mirror (in which the results deteriorate significantly with the noise), the behaviour is very similar to what happens in the previous evaluation.

\subsection{Relative Rotation in a Simulated Environment} \label{sec:relative_pose}

\begin{figure}
  \vspace{-0.3cm}
  \includegraphics[height=4.1cm]{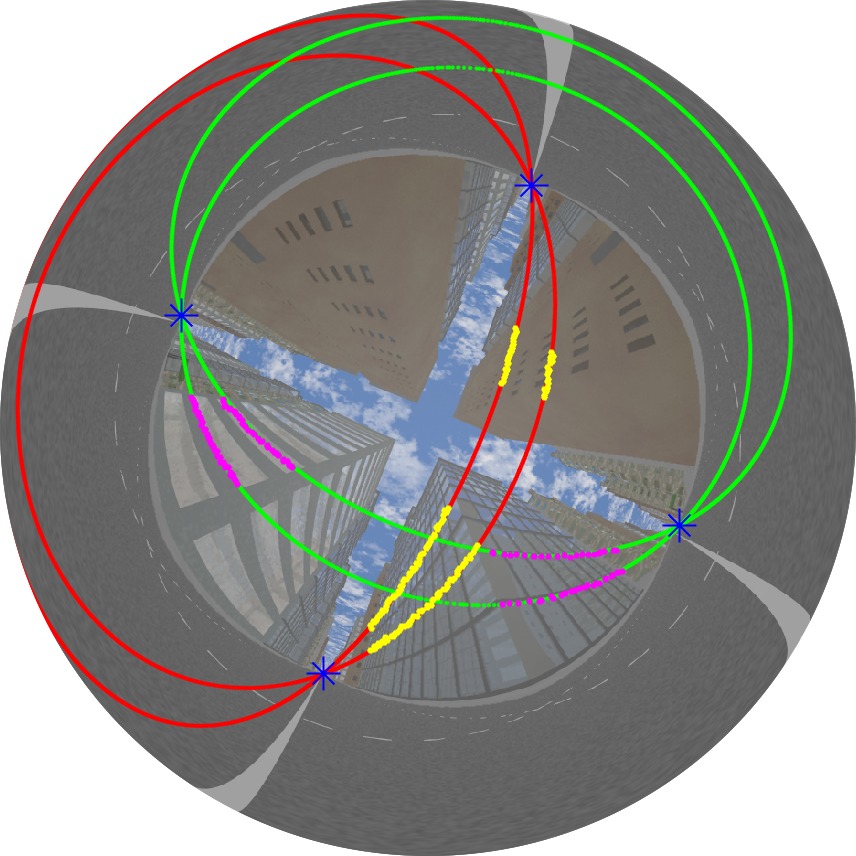}
  \includegraphics[height=4.1cm]{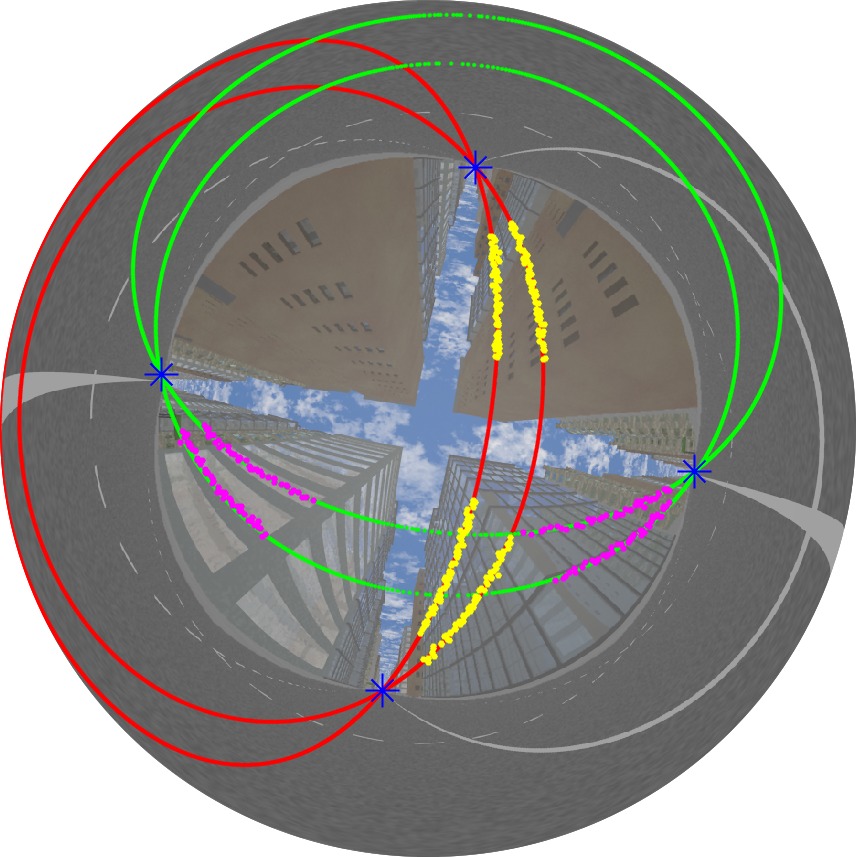}
  \caption{Images used to get the camera orientation from a sequence of two images. Curves are fitted using the yellow and magenta pixels (with a noise standard deviation of three pixels), and we match projection curves from parallel lines using the same colors in both images. We compute vanishing points from the intersection of the curves (blue points in both images), and get the correspondent directions. Then, the rotation is computed using the {\it Procrustes} problem.}
  \label{fig:relative_pose_sims}
\end{figure}

This subsection address the estimation of the rotation parameters, in a relative pose problem. This problem can be solved using vanishing points and a method similar to what was proposed in Sec.~\ref{sec:experiments_estimation_rotation_parameters}, assuming as known a matching of $N \geq 2$ vanishing points in a couple of images. To generate these images, we use {\tt POV-Ray} \cite{povray}. A non-central ellipsoidal catadioptric camera system was simulated in an urban environment. We acquire images from two distinct points of
view (as shown in Fig.~\ref{fig:relative_pose_sims}).

To estimate rotation we consider the following procedure: 1) we fit parallel curves on both images considering noise in the image points of two pixels (standard deviation), see the yellow and pink points in Fig.~\ref{fig:relative_pose_sims}; 2) compute corresponding vanishing points in both images; 3) compute directions using the method presented in Sec.~\ref{sec:compute_direction_vanishing_points}; and 4) estimate the rotation that aligns both camera coordinate system using the Procrustes' problem. The difference between the estimated rotation matrix and the ground truth was: $0.0446$, which agrees with the evaluation shown in Sec.~\ref{sec:experiments_estimation_rotation_parameters}.

\begin{figure}
\vspace{-0.3cm}
  \subfloat[Spherical catadioptric camera configuration.]{
    \includegraphics[width=0.23\textwidth]{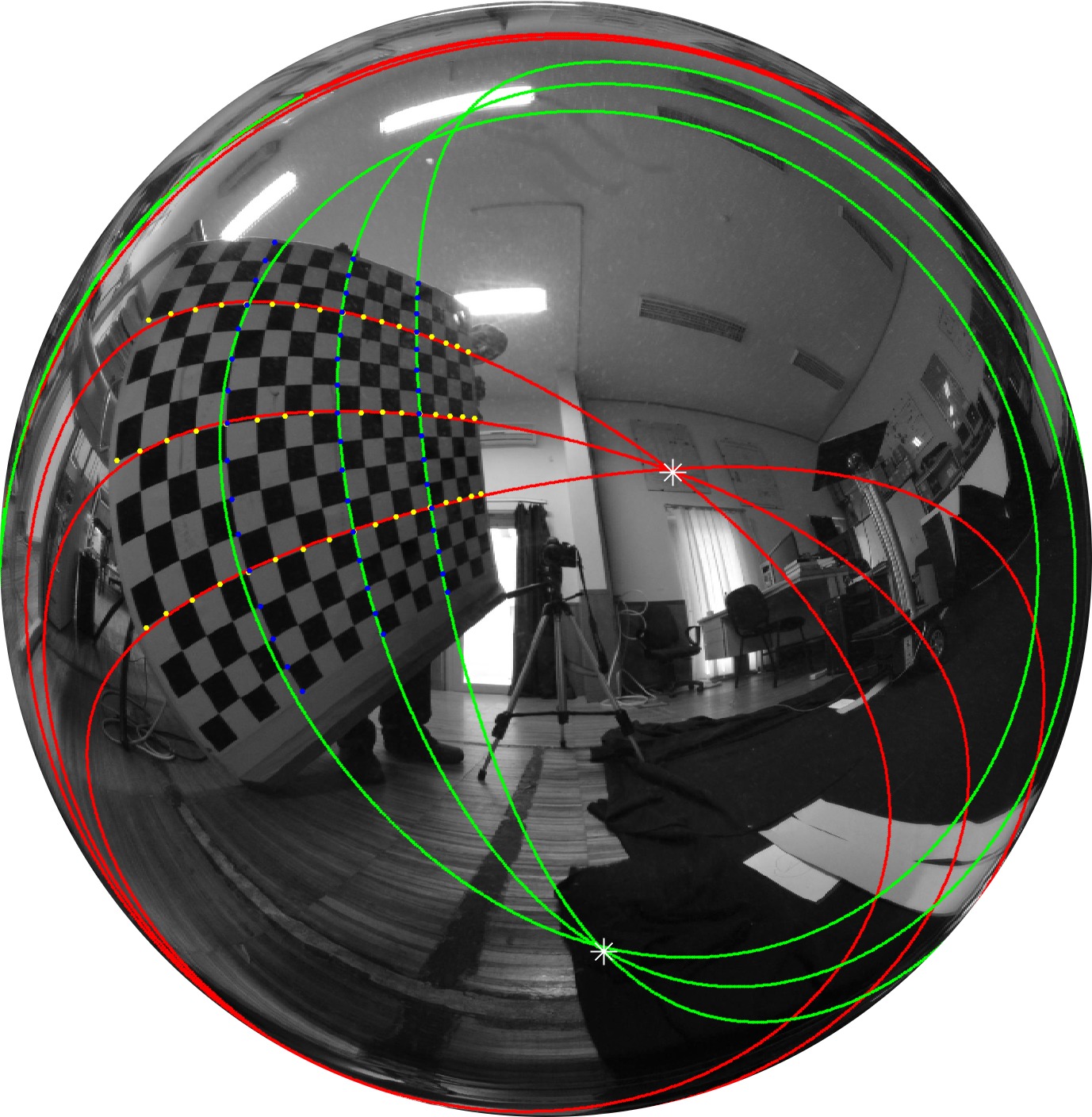} \hfill
    \includegraphics[width=0.23\textwidth]{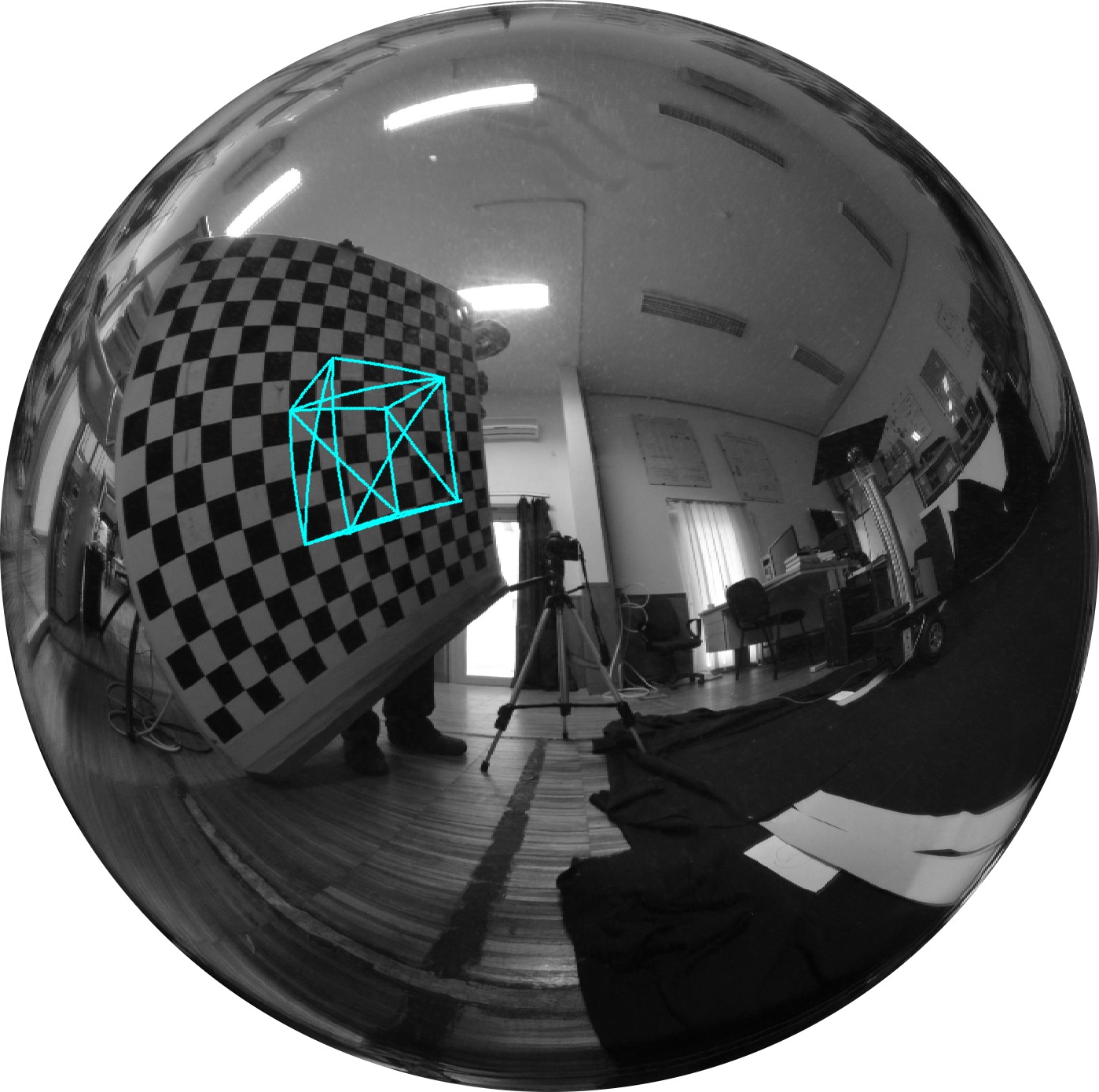}
  }\newline
  \subfloat[Hyperbolic catadioptric camera configuration.]{
    \includegraphics[width=0.23\textwidth]{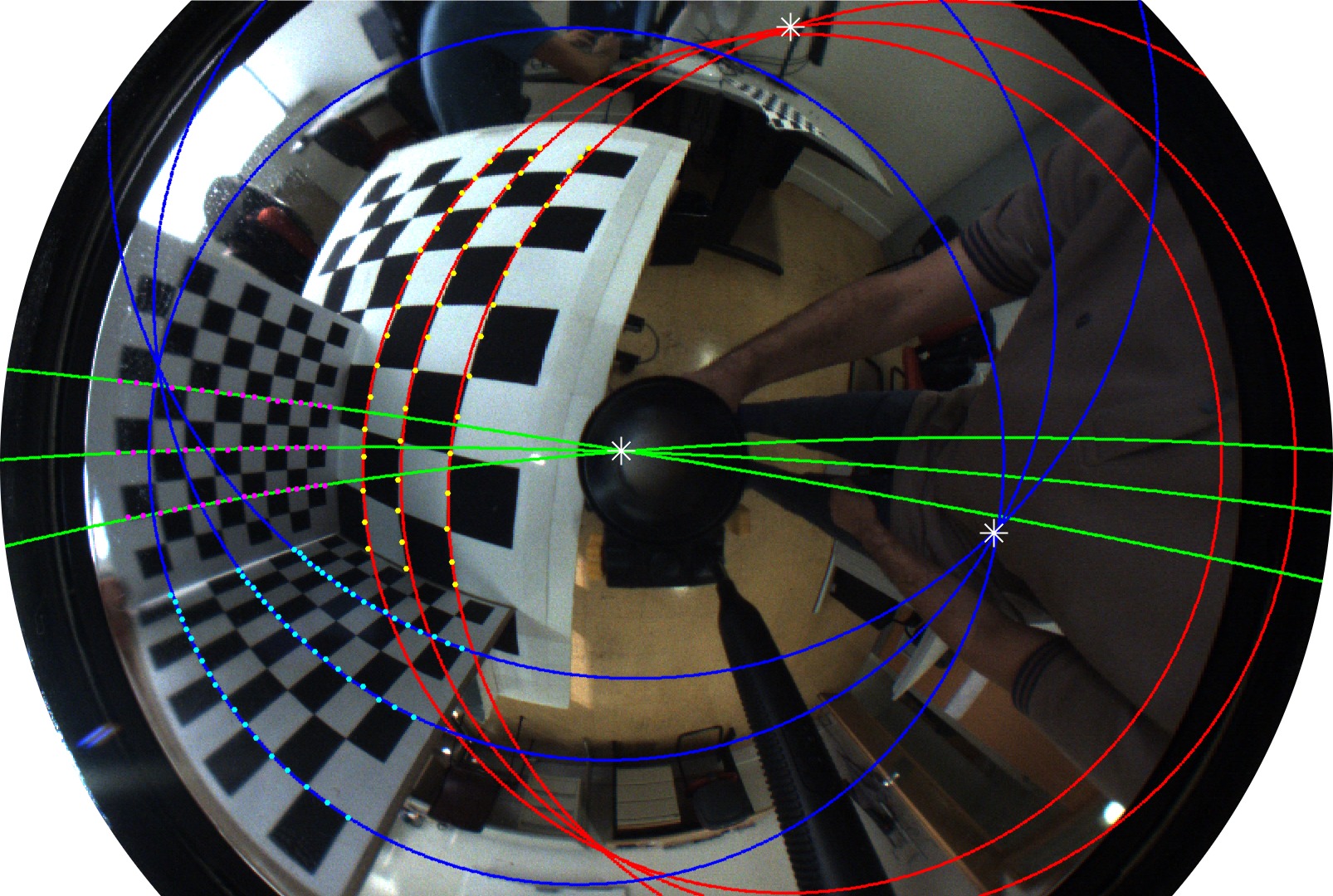} \hfill
    \includegraphics[width=0.23\textwidth]{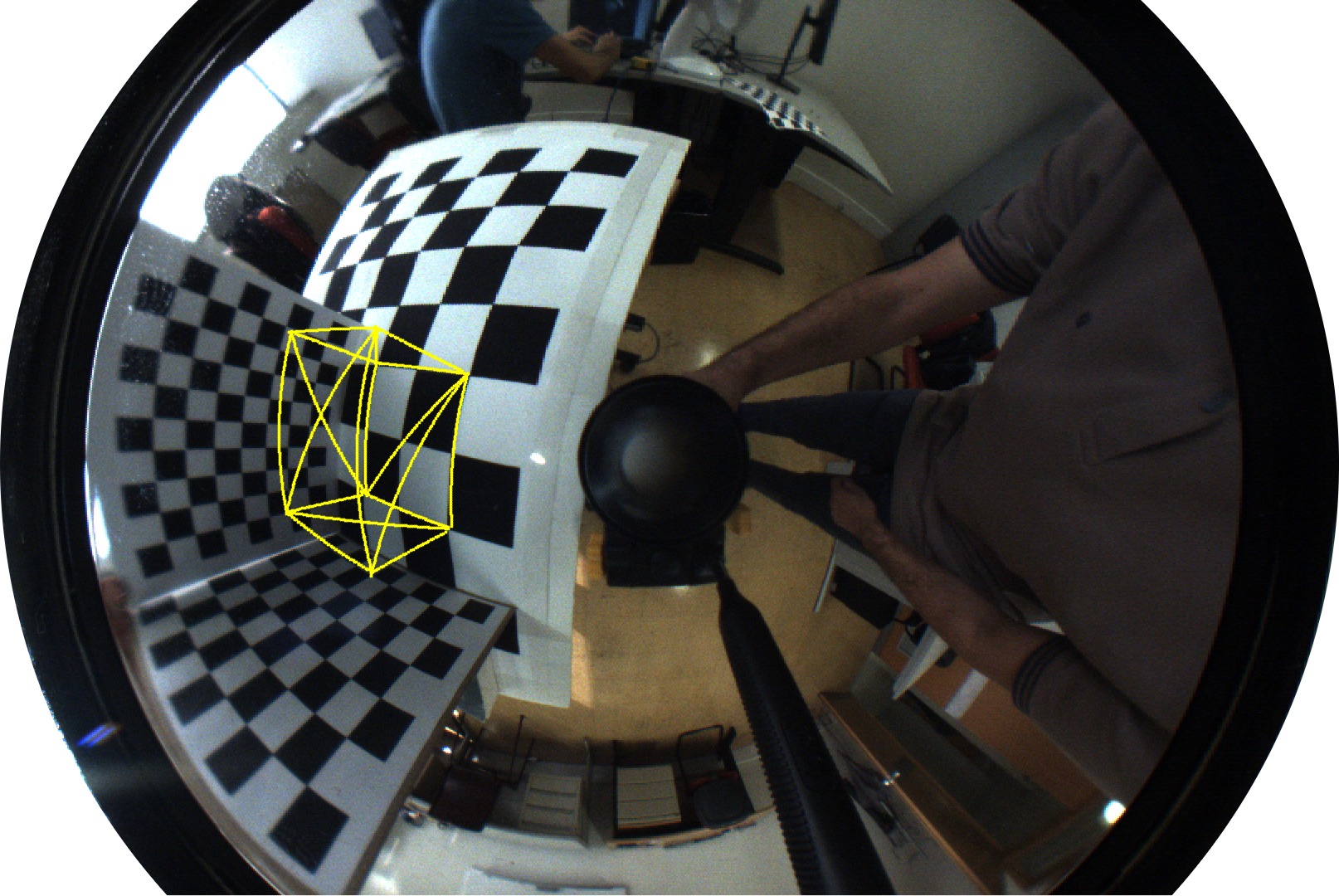}
  }
  \caption{Application of the proposed solution for the absolute pose problem, using both hyperbolic and spherical mirrors. On the left we show the fitting of curves in the image (using the colored points on the chessboard). Then, we compute vanishing points from the intersection of the curves with the same color, the respective directions from vanishing points, and the rotation and translation parameters. To validate the camera pose, we define a square on the chessboard (3D world) and project it on the image. The cube is drawn in the correct position (image space), which validates the computation of the camera pose.
    }
  \label{fig:real_pose}
\end{figure}

Fitting curves in non-central catadioptric cameras involves the estimation of the coefficients of high degree polynomial equations, which are hard to recover robustly. In this paper, we consider the following procedure: 1) find an initial estimate (non-robust solution) to the 3D parameters using one of the state-of-the-art techniques; 2) fit the curves in the mirror; 3) compute the distance between the curves and the pixels that were supposed to be on the associated lines; 4) update the 3D line parameters and iterate these 4 steps till the distance is less than a threshold.

\subsection{Absolute Pose using Real Data} \label{sec:real_data}
We consider two real non-central catadioptric camera systems using a spherical and a hyperbolic mirror. The perspective cameras were calibrated using the {\tt Matlab} toolbox and we used the mirror parameters specified by the manufacturer. The transformation between the mirror and perspective cameras was computed using \cite{perdigoto13}.

We captured chessboard images and used their corners to compute parallel lines. We consider two scenarios, one of them in which each chessboard is used to find vanishing points from a basis of the world coordinate system; and one with a single chessboard, in which we can identify two vanishing points that are two perpendicular directions in the world (i.e. the minimal case). In addition, we assume that the measurements of one of the chessboards is known, i.e. we know the coordinates of some 3D line on that chessboard and their correspondent images (pixels). We fit lines using the method presented in the previous subsection, and compute the respective vanishing points, see the results in Fig.~\ref{fig:real_pose}. Then, we compute the pose using the method described in Sec.~\ref{sec:app_camera_pose}. After that, we define a cube in the chessboard coordinates system, and project its wireframe into the image using \cite{agrawal11,dias15}. Fig.~\ref{fig:real_pose} proves the pose was computed correctly.

\begin{figure}[t]
\vspace{-0.3cm}
\begin{center}
   \subfloat[]{\includegraphics[width=0.38\linewidth]{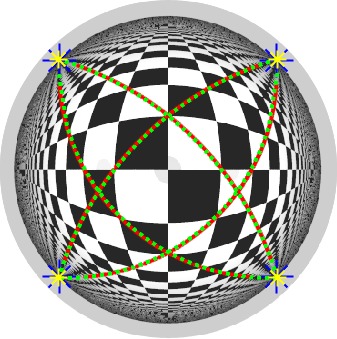}\label{a}} \quad
   \subfloat[]{\includegraphics[width=0.55\linewidth]{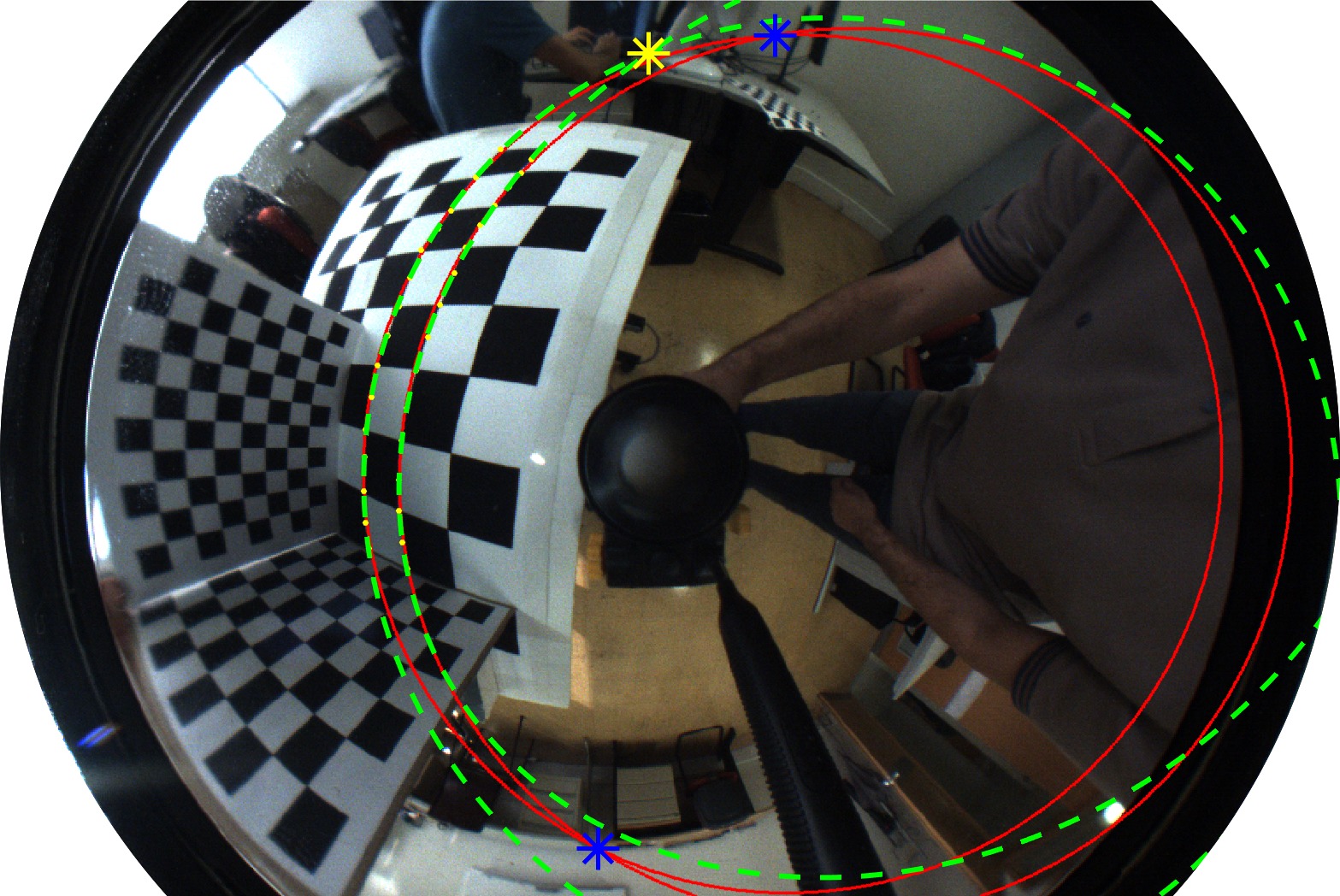}\label{b}}
\end{center}
   \caption{Comparison between the vanishing points using our method (red curves and blue points), and the approximation using the unified central model (green curves and yellow points). We show examples of POV-Ray \protect\subref{a} (to get a central omnidirectional camera) and real images \protect\subref{b}, in which the camera is about $9\%$ away from the central configuration. Regarding \protect\subref{b} we got a deviation of $17^{\circ}$ between our exact direction and the one given by the unified model.}
\label{fig:long}
\label{fig:onecol}
\end{figure}

\subsection{Modeling Central and Non-Central Solutions}
\label{sec:central_vs_ncentral}
We compare the vanishing points using the general approach with the one using central unified model (presented in the supplementary material). As expected, in the central case, the general approach gets the same solution as the unified model, as shown in Fig.~\ref{fig:onecol}\subref{a}. In Fig.~\ref{fig:onecol}\subref{b} we test a non-central configuration with a hyperbolic mirror (the camera is about $9\%$ away from the central configuration). We approximate the camera using the unified camera model, and plot the solutions using the derived fourth degree polynomial, drawing also the results based on the general theory derived in this paper. The considerable improvement shown by the general approach over the central unified model asserts that our theory for general catadioptric cameras is essential for accurate modeling in non-central cameras.

\begin{figure}
  \centering
  \includegraphics[height=3.2cm]{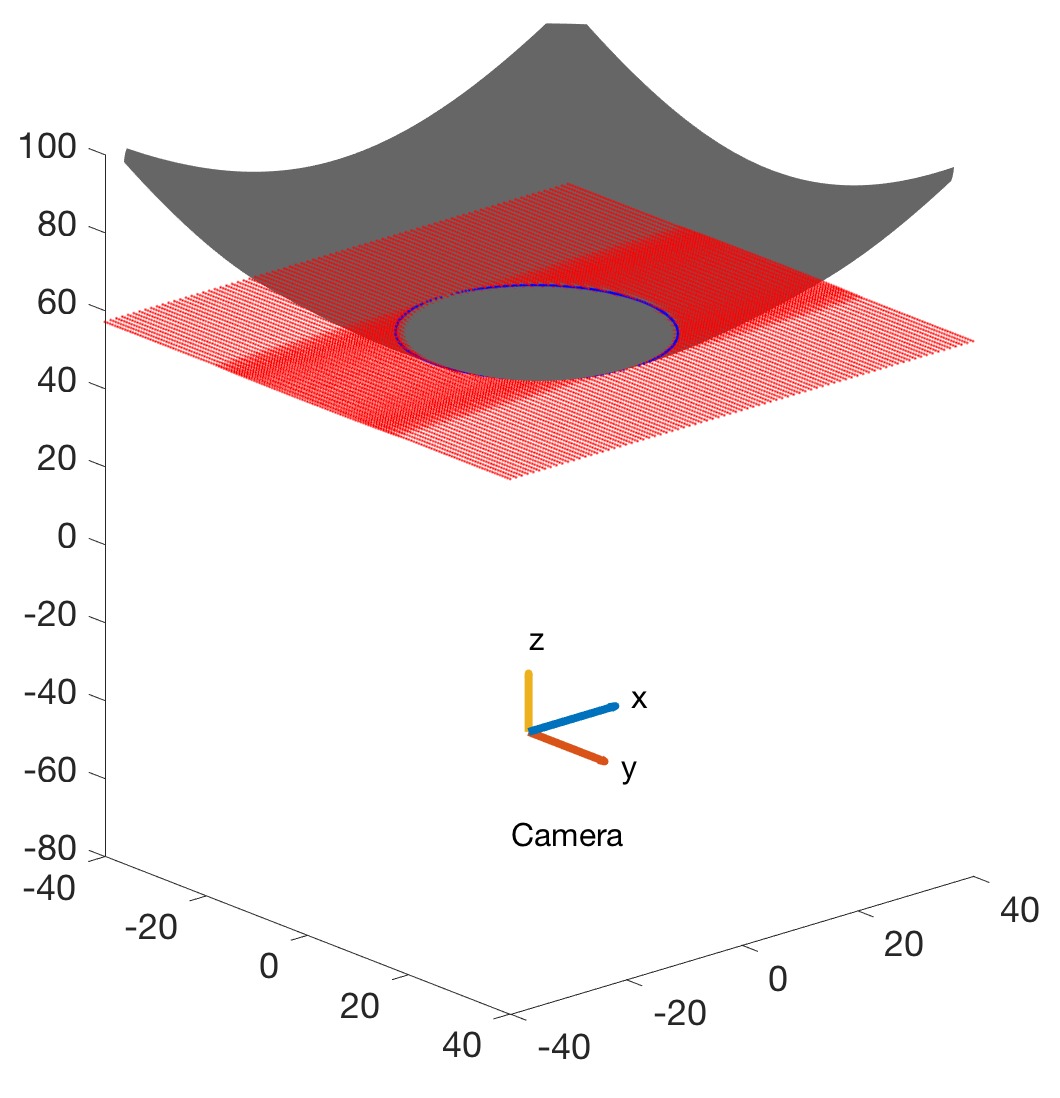} \hfill
  \includegraphics[height=3.2cm]{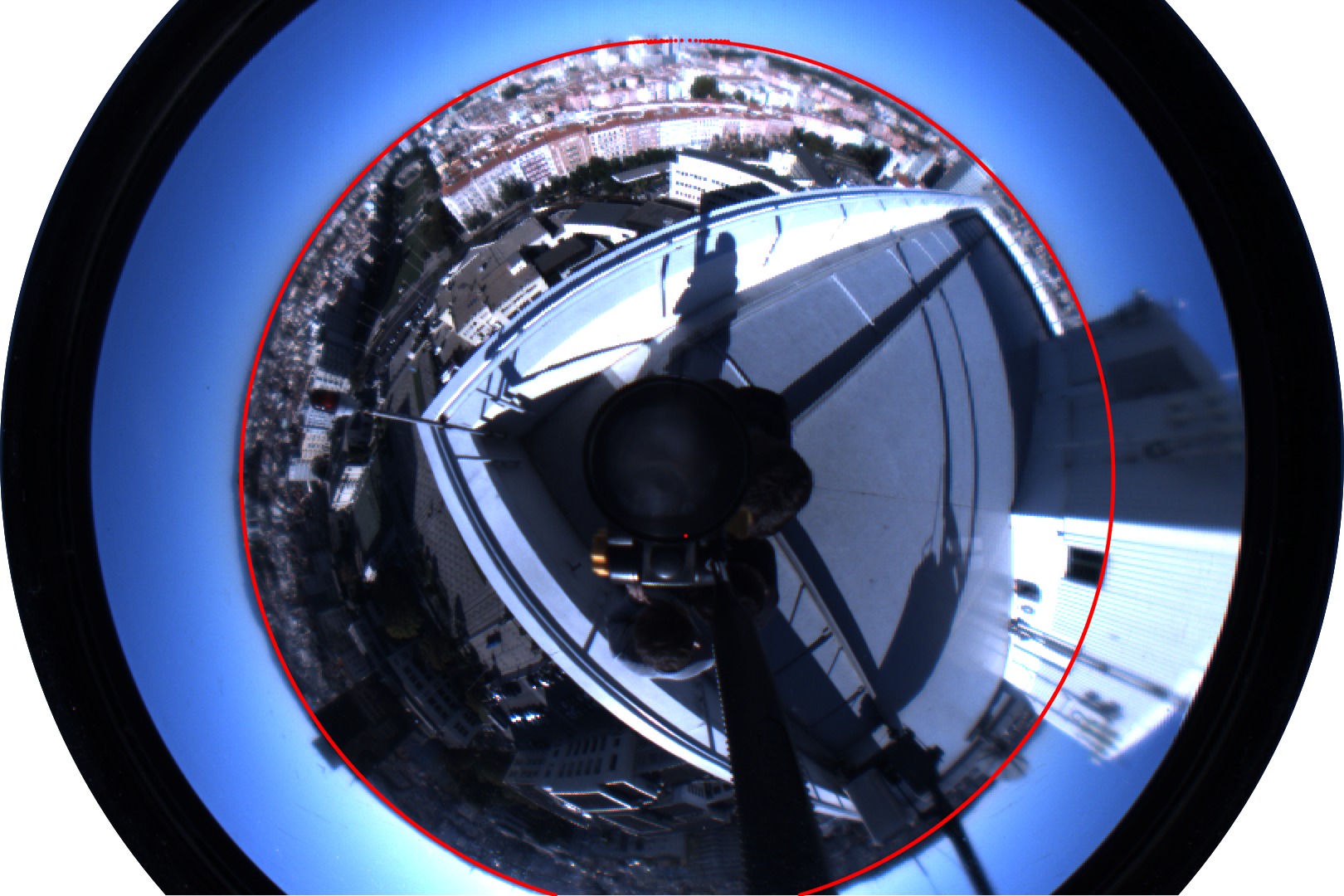}
  \caption{An image was taken from the top of a building. We measure the rotation of the camera w.r.t. the ground plane, and use this information to define its normal vector. With this vector, we apply the method derived in Sec.~\ref{sec:curves_infinity} and draw the curve (in red) in the image that can be used for partitioning it in to two distinct regions, namely the ground and the sky. At the left, we show the intersection of both surfaces that define the curve: 1) $\Gamma(\mathbf{r})$ in red; 2) and $\Omega(\mathbf{r})$ gray surface representing the mirror.
  } 
  \label{fig:vanishing_lines}
\end{figure}
\subsection{Vanishing Curves}\label{sec:exp_vanishing_lines}
We show vanishing curve estimation using a non-central hyperbolic catadioptric camera. For that purpose, we consider a camera system used in the previous experiments, and took a picture outside, where the horizon line was visible. We specifically choose a camera position almost perpendicular to the ground floor ($\mathbf{n} = [0.05\text{ }0.05\text{ }0.997]$), which allows us to extract directly the parameterization of the curve as derived in Sec.~\ref{sec:curves_infinity}. The results are shown in Fig.~\ref{fig:vanishing_lines}, in which we show the intersection of both surfaces ($\Gamma(\mathbf{r})$ and $\Omega(\mathbf{r})$), and the resulting curve in the image. Qualitative validation of the vanishing curve estimation is achieved from the precise alignment of the curve with the horizon. 

\section{Discussion}

We propose analytical modeling of vanishing points and vanishing curves in general omnidirectional cameras. To the best of our knowledge, there is no prior work on parametric modeling of vanishing points and vanishing curves for a taxonomy of catadioptric cameras. We propose solutions for vanishing point estimation from line directions, and line direction estimation from vanishing points. The proposed methods differ in terms of computational complexity. The computation of a vanishing point given a direction of a 3D line consists in solving for the roots of one polynomial of degree at most 10 and two back substitutions, as presented in Sec.~\ref{sec:compute_vanishing_points}. The problem of computing a direction from a vanishing point corresponds to three polynomial evaluations described in Sec.~\ref{sec:compute_direction_vanishing_points}, which can be computed analytically. The proposed methods are evaluated in both simulations with noise and real data. In future, we plan to use the estimated vanishing points and vanishing curves in the context of large-scale line-based 3D modeling of Manhattan scenes using catadioptric cameras.

\section*{Acknowledgments}
P. Miraldo is with the Institute for Systems and Robotics (ISR/IST), LARSyS, Instituto Superior T\'{e}cnico, Univ Lisboa. F. Eiras was with ISR/IST and he is now with Linacre College of the University of Oxford. S. Ramalingam acknowledges the support from Mitsubishi Electric Research Labs (MERL). This work was partially supported by the Portuguese projects [UID/EEA/50009/2013] \& [PTDC/EEI-SII/4698/2014] and grant [SFRH/BPD/111495/2015]. We thank the reviewers and area chairs for valuable feedback.  
{\small
\bibliographystyle{ieee}
\bibliography{ref}
}

\end{document}